# Mixed-Reality-Guided Teleoperation of a Collaborative Robot for Surgical Procedures

Gabriela Rus[1][0000-0002-1751-8952], Nadim Al Hajjar[3][0000-0001-5986-1233], Paul Tucan[1][0000-0001-5660-8259], Andra Ciocan[3][0000-0003-0126-6428], Calin Vaida[1][0000-0003-2822-9790], Corina Radu[4][0000-0003-0005-0262], Damien Chablat [1,5][0000-0001-7847-6162]
Doina Pisla[1,2][0000-0001-7014-9431]*

[1]CESTER, Technical University of Cluj-Napoca, 400114 Cluj-Napoca, Romania
[2]Technical Sciences Academy of Romania, 26 Dacia Blvd, 030167 - Bucharest, Romania
[3]Department of Surgery, "Iuliu Hatieganu" University of Medicine and Pharmacy, 400347 Cluj-Napoca, Romania
[4] Department of Internal Medicine, "Iuliu Hatieganu" University of Medicine and Pharmacy, 400347 Cluj-Napoca, Romania
[5] École Centrale Nantes, Nantes Université, CNRS, LS2N, UMR 6004, F-44000 Nantes, France
*Corresponding author

**Abstract.** The development of advanced surgical systems embedding the Master-Slave control strategy introduced the possibility of remote interaction between the surgeon and the patient, also known as teleoperation.
The present paper aims to integrate innovative technologies into the teleoperation process to enhance workflow during surgeries. The proposed system incorporates a collaborative robot, Kuka IIWA LBR, and Hololens 2 (an augmented reality device), allowing the user to control the robot in an expansive environment that integrates actual (real data) with additional digital information imported via Hololens 2. Experimental data demonstrate the user's ability to control the Kuka IIWA using various gestures to position it with respect to real or digital objects. Thus, this system offers a novel solution to manipulate robots used in surgeries in a more intuitive manner, contributing to the reduction of the learning curve for surgeons. Calibration and testing in multiple scenarios demonstrate the efficiency of the system in providing seamless movements.

## 1 Introduction

The concept of teleoperation in surgery marks a significant advancement in medical technology, allowing surgeons to manipulate instruments from a remote location with precision. This innovative approach enables the execution of intricate surgical procedures through robotic systems [1], enhancing the capabilities of medical professionals, especially for master-slave systems [2][3].

According to this, it is important to underline how critical are the aspects related to precision and guidance. Surgeons need high precision to achieve goals while guidance offers real-time information for informed decisions. As medical professionals continually seek innovative solutions to improve patient outcomes, the integration of cutting-edge technologies becomes primary [4] [5] [6] [7]. A significant component in this effort is the existent collaborative robotics systems.

The advantages of using collaborative robots in surgical procedure include (1) safety futures-force and torque sensors, vision systems, and software algorithms that allow them to detect the presence of humans and respond by slowing down or stopping; (2) adaptability and flexibility – due to their configuration can be easily reconfigured for various surgical procedures; (3) human-robot collaboration - these system being designed to interact with humans. Considering the presented aspects these systems showed their potential to be used in various complex surgery procedures such as Esophagectomy (esophageal resection), Pancreatectomy (pancreatic resection) and Hepatectomy (liver resection), which share a common issue, referring to the location of the tumors in the proximity of critical internal structures imposing highly accurate resection techniques.

The use of collaborative robots in surgery was discussed in many articles [8-11]. For example, the authors from [10] used the UR5 robot to hold and move the surgical instruments and the camera during surgeries. The authors of [11] used KUKA LWR 4+ robot to develop a collaborative control framework for teleoperations in Minimally Invasive Surgery (MIS).

Another innovative technology increasingly used in recent years, demonstrating real advantages in surgery is related to the augmented reality (AR) and virtual reality (VR) devices. These devices have the potential to improve surgeries in many aspects such as: surgical planning and visualization, training and simulation, enhanced navigation, remote assistance, reduced radiation exposure and postoperative monitoring and rehabilitation [12][13].

The paper aims to analyze the potential of utilizing collaborative robots and an AR device (Hololens 2) to enhance teleoperation. The proposed system is designed to enable the manipulation of surgical robots without the need for additional joysticks such as 3D space-mouses or haptic devices, offering an intuitive interface. Simultaneously, it provides the capability for multi-user operation of the robot, allowing multiple users with HoloLens devices to use the system. Considering these aspects, the system can be used in intraoperative phase, having the potential to reduce learning curves and enhance the overall surgical process for surgeons.

Succeeding the introduction section, the paper is structured as follows: Section II describes the methodology for development of the proposed system including calibration and the protocol for testing. Section III describes the results, and Section IV presents the discussions and the conclusions of the study.

## 2      Methodology

The teleoperation setup proposed for this study is composed of a collaborative robot, a laparoscopic camera attached to the robot through a customized flange, and the augmented reality device, HoloLens 2, as can be seen in Fig.1. The transfer of data between systems is based on TCP/IP (Transmission Control Protocol/Internet Protocol) communication.

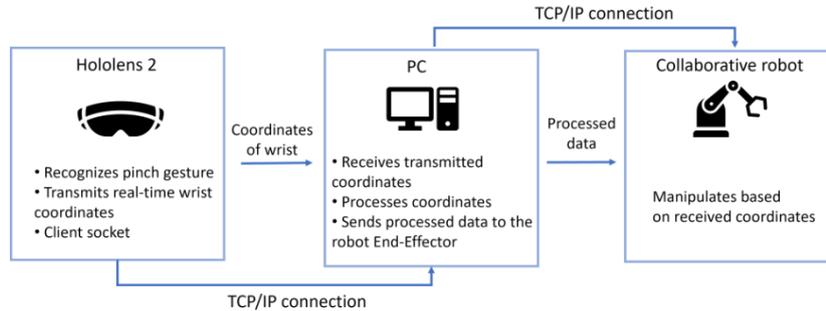

**Fig.1** Schematic representation of the proposed system

### 2.1 Materials

The chosen collaborative robot was **KUKA IIWA** (Intelligent Industrial Work Assistant) [14]. The configuration of the system (seven joints) includes integrated force torque sensors for safe and efficient operation, which enables suitable control of movement. Its collaborative features not only ensure safety but also make it easy for tasks which demand precise manipulation. A laparoscopic camera is connected to a specially tailored flange, this assembly being fixed to the end effector of the robot.

**HoloLens 2,** a Microsoft augmented reality device, introduces a hands-free advantage to the described setup. The HoloLens 2 hands free control makes it easier to perform precise movements based on hand movement, as part of teleoperation set-up.

### 2.2 Communication

As can be seen in Fig.2 the communication is based on TCP/IP connection, assuring real-time transfer of data.

In the described teleoperation framework, the HoloLens device hosts a standalone application that provides several functions including a client socket. This application is designed to transmit real-time wrist coordinates when a certain gesture (the pinching) is recognized, these coordinates being then directed to a server socket, created in Python. Thus, upon initiating the application, it verifies the connection with the Python server, and the application proceeds only when the connection between the client and server is established. The same prerequisite applies to the connection between Python and the robot. If the connection is successfully established among all participants, the pinch gesture from HoloLens is verified, and the data transfer is initiated upon the recognition of the gesture. The coordinates are further processed and sent to the end effector of the KUKA robot after being received by the Python server, enabling real-time control and manipulation. After the pinch gesture is finished, HoloLens does not transmit any more coordinates until the gesture is performed again, which stops unintentional robot movements. In order to assure the transfer of all of required data for a movement, a validation of the movement is implemented in the user interface, thus the user will validate if the movement was executed by the robot or not.

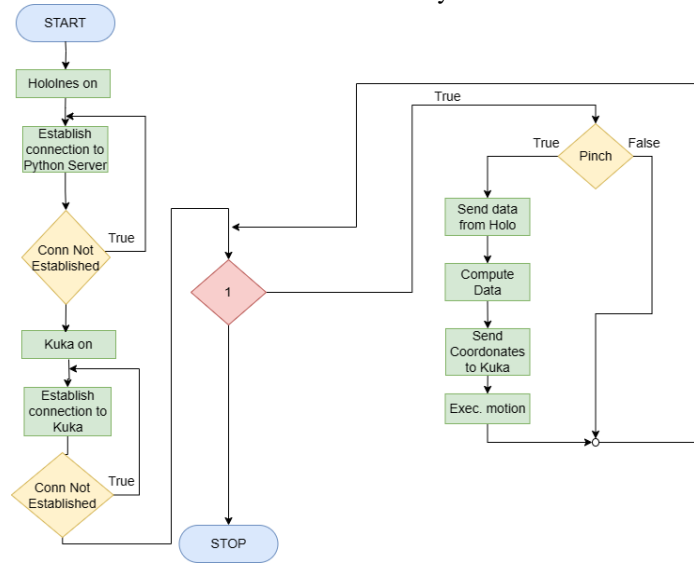

**Fig. 2.** Schematic representation of the data transfer between Hololens and Kuka

### 2.3 Calibration

Hololens has been configured to consider the KUKA coordinate system throughout the calibration process, ensuring seamless integration and accurate interpretation of coordinates between the augmented reality device and the collaborative robot.

To address the unit disparity between the KUKA system, which operates in millimeters, and the Hololens, which uses meters, a normalization step was implemented.

Additionally, as coordinates are transmitted to the end effector of the KUKA collaborative robot, the calibration process incorporates the utilization of inverse kinematics.

For this procedure, the coordinates from the End-Effector are required. The origin of the coordinate system is positioned in the characteristic point of the End-Effector, and the axes aligned with the robot native ones enabling a fast computation with respect to the base of the robot. The integration of robot singularities and trajectory planning has been realized using the Kuka Sunrise Toolbox.

As part of the calibration process, the robot was moved to a known point in space. The coordinates of the tip of the laparoscopic camera were then registered for comparison using specific markers with locations defined accurately with respect to the fixed coordinates system of the robot. A similar process was carried out for the cursor from HoloLens.

Ultimately, the robot is positioned in a specific configuration, serving as the starting point for task execution. This strategic stance facilitates the seamless initiation of various tasks.

### 2.4 Protocol for testing

The protocol for testing and ensuring the functionality of the collaborative robot, integrated with the Hololens, involves several key steps presented in Table 1. In the initial setup, the correct configuration and connection of the robot, as well as verifying the proper functioning of the Hololens is performed. Basic movement tests are then conducted in an open space to guarantee accurate motion. Calibration between the robot and Hololens is tested, and the recognition of pinch gestures with different hand orientations, is verified. Subsequently, specific movement tests are executed, with emphasis on the insertion of the laparoscopic camera through the trocar. The laparoscopic camera has 3 degrees of freedom (DOF) and can perform 2 rotational movements around the X-axis, and respectively around the Y-axis, as well as translation along the Z-axis (Fig 3.a).

**Table 1.** Testing protocol.

| Steps of protocol | Description |
| --- | --- |
| **Initial setup** | • Ensure proper installation and connection of the collaborative robot<br>• Ensure Hololens functionality |
| **Basic Movements Tests** | • Perform basic movement tests in open space to ensure the robotic moves smoothly and accurately.<br>• Test the calibration between the robot and Hololens<br>• Test the recognition of the gesture – with pinch gesture, without pinch gesture, with different orientation of the hand |
| **Task 1** | **Move laparoscopic camera on plane** (The movements are performed on a 2D plane). **Task description**: A paper where diverse geometrical figures are drawn, is placed in front of Kuka. The robot, based on the hand's movement, should be able to move to a certain point of a figure or to execute movements throughout the figures from the paper. |
| **Task 2** | **Move laparoscopic camera on 3D space**. (The movement is performed on the 3D space). **Task description**: In proximity of the robot are placed both physical objects (a phantom torso) and holograms (projected from Hololens) The robot should be able to gentle touch these holograms, respectively to introduce the laparoscopic camera in torso. |
| **Task 3** | **Following a trajectory**. **Task description:** The robot should be able to move based on certain trajectory drawn by the hand in order to enhance maneuverability and accuracy. |

## 3 Results

Considering that the main purpose of this paper is to test the feasibility of a system composed from an AR device and a collaborative robot, in teleoperation surgery, aspects such as accuracy, delays and time to complete a task are critical metrics. The accuracy was calculated taking into account the coordinates of the cursor from Hololens and the coordinate of the tip of laparoscopic camera (coordinates of the End-Effector + the offset caused by the length of the camera – 30 cm). The average positional error was measured at 0.7 mm during the execution of movement 1, presented in protocol. The positional errors registered for the next step in protocol (movement 2), were 0.8 mm when the user had to touch the holograms with the laparoscopic camera and 1.1 mm when the task was the introduction of laparoscopic camera in torso (Fig 3.b).

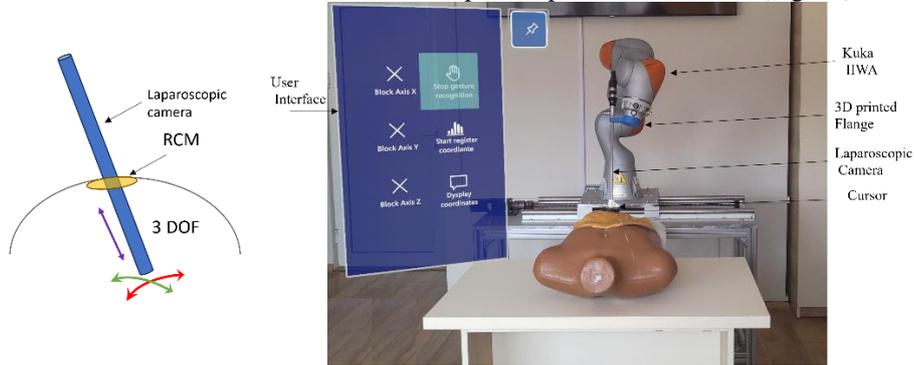

**Fig 3. a)** Motions of camera

**Fig 3. b) Execution of task 2.** Inserting the laparoscopic camera in the phantom torso (image from Hololens)

For the final test the positions of the end effector and the positions of the hand have been registered to analyze the trajectories. As can be seen in Fig.4, the collaborative robot has been able to follow closely the trajectory of the hand.

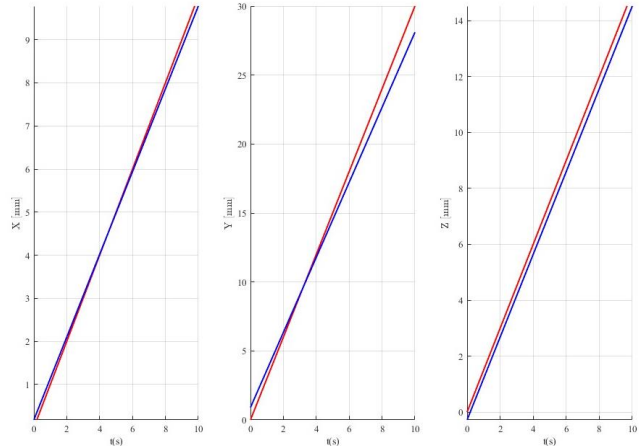

**Fig 4.** Graphical representation of the trajectory registered from the wrist of the hand (red line) and End-Effector trajectory (blue line)

Considering the protocol for data transmission, the average delay registered between the hand and the robot was almost 0.1 milliseconds. It has been observed that sudden movements tend to create an additional delay caused by the large amount of data that should be processed.

The average time to complete the tasks was 1 minute and 45 seconds for the first one, 2 minutes and 15 seconds for touching all holograms, 1 minute and 52 seconds for introducing the camera in the torso, and 40 seconds for task 3. These tasks were completed by three users with different levels of experience with the Hololens device.

## 4      Discussion and conclusion

Based on the tests performed with the proposed system, Kuka collaborative robot demonstrated the capability to effectuate accurate movements based on real time commands from Hololens, for all proposed tasks. However, tasks of varying complexity showed proportional errors, as can be seen between Task 1 and Task 2, reflecting the demand for increased dexterity. Data transmission was executed with a small delay, which indicates the potential of this system to be used in real-time applications such as surgery.

The proposed system is feasible for teleoperation, bringing in addition the advantage of allowing multi-user manipulation if there are more Hololens devices, facilitating collaborative decision-making and enhancing overall surgical efficiency. This real-time, intuitive control system has the potential to optimize surgical procedures, reduce learning curves, and improve outcomes for the surgical team.

Another advantage is given by its collaborative design, thanks to which potential collisions do not lead to damage. Additionally, the robot provides the option to restrict each joint within a virtual plane, ensuring that the robot stays within predefined spatial boundaries.

To respond to the specific task of holder for laparoscopic camera, there are other aspects that should be also taken into consideration. First of all, in order to avoid unwanted behavior, the robot executes a linear displacement to the insertion point. Before the insertion, the camera is aligned with the insertion point. Following this alignment, the insertion is carried out incrementally, with minimal increment and minimal velocity (both the velocity and the increment are selected by the user from the interface). Another important factor is the scaling between the movement of the human operator and the robot. This component is also implemented in the user interface and allows the user to scale the movement, in accordance with his needs.

In conclusion, the teleoperation setup combining Kuka IIWA and HoloLens 2 can bring a significant advancement in surgical technology. The collaborative system, together with HoloLens device, used in intraoperative phase holds promise for improving surgical precision, safety, and overall outcomes.
This system, built around the collaboratives robot and the augmented reality device HoloLens 2, can be easily integrated into diverse collaborative systems.

### Acknowledgment

This work was supported by the project New smart and adaptive robotics solutions for personalized minimally invasive surgery in cancer treatment - ATHENA, funded by European Union – NextGenerationEU and Romanian Government, under National Recovery and Resilience Plan for Romania, contract no. 760072/23.05.2023, code CF 116/15.11.2022, through the Romanian Ministry of Research, Innovation and Digitalization, within Component 9, investment I8 and  by a grant of the Ministry of Research, Innovation and Digitization, CNCS/CCCDI—UEFISCDI, project number PN-III-P2-2.1-PED-2021-2790 694PED—Enhance, within PNCDI III.